\documentclass[11pt]{article} 

\usepackage{geometry} 
\usepackage{graphicx} 
\usepackage{multirow}
\usepackage{amsmath}
\usepackage{amssymb}
\usepackage{placeins}

\usepackage{booktabs} 
\usepackage{array} 
\usepackage{paralist} 
\usepackage{verbatim} 

\usepackage{fancyhdr} 
\usepackage{tikz}
\usetikzlibrary{shapes,arrows}
\usepackage{caption}

\usepackage{sectsty}
\allsectionsfont{\sffamily\mdseries\upshape} 

\usepackage[nottoc,notlof,notlot]{tocbibind} 
\usepackage{amsmath,amssymb}

\usepackage{changepage}

\usepackage[utf8]{inputenc}

\usepackage{textcomp,marvosym}

\usepackage{cite}

\usepackage{nameref,hyperref}

\usepackage[right]{lineno}

\usepackage{microtype}
\DisableLigatures[f]{encoding = *, family = * }

\usepackage{array}

\usepackage{subcaption}
\usepackage{amssymb}
\usepackage{amsmath}
\usepackage{amsthm}
\usepackage{fixmath}
\usepackage{lipsum}
\usepackage{caption}
\usepackage{verbatim}
\usepackage{float}
\usepackage{tikz}

\newcolumntype{+}{!{\vrule width 2pt}}

\newlength\savedwidth

\usepackage{titlesec}
\titleformat{\paragraph}
{\normalfont\normalsize\bfseries}{\theparagraph}{1em}{}
\titlespacing*{\paragraph}
{0pt}{3.25ex plus 1ex minus .2ex}{1.5ex plus .2ex}

\usepackage[aboveskip=1pt,labelfont=bf,labelsep=period,justification=raggedright,singlelinecheck=off]{caption}

\makeatletter
\renewcommand{\@biblabel}[1]{\quad#1.}
\makeatother

\newcommand{\experiment}[2]{%
    \vspace{0.8em}
    \noindent\textbf{#1: #2.}\par
    \vspace{0.2em}
}

\newlength{\vpcpanelheight}
\theoremstyle{definition}

\providecommand{\keywords}[1]
{
  \small	
  \textbf{\textit{Keywords---}}#1
}

\begin{document}

\vspace*{0.2in}

\begin{flushleft}
{\Large
\textbf\newline{Multimodal Empirical Bayes Variational Autoencoders for Joint Longitudinal and Time-to-Event Modeling} 
}
\newline

Anders Sjöberg\textsuperscript{1,2}*\textsuperscript{†},
Nils Olsson\textsuperscript {1}*,
Marcus Baaz\textsuperscript{1}*,
Mats Jirstrand\textsuperscript{1,2}

\bigskip
\textbf{1} Fraunhofer-Chalmers Centre, Gothenburg, SE-412 88, Sweden
\\
\textbf{2} Department of Electrical Engineering, Chalmers University of Technology, Gothenburg, SE-412 96, Sweden
\\
* Equal Contribution
\\ 
† Corresponding author: andsjob@chalmers.se
\bigskip

\end{flushleft}

\begin{abstract}
Longitudinal tumor measurements, dropout information, and genetic covariates provide complementary information about treatment response, but integrating these data sources within a single population modeling framework remains challenging. We extend the empirical Bayes variational autoencoder (EB-VAE) framework to joint longitudinal and time-to-event modeling and evaluate it on tumor growth data. The framework represents inter-individual variability using latent individual effects regularized by a covariate-conditioned empirical Bayes prior, while a decoder maps these latent effects to tumor-volume trajectories. To account for informative dropout, the decoder was augmented with a hazard model, yielding joint predictions of tumor growth and time to dropout. We further compared fully neural and hybrid semi-mechanistic decoder formulations and incorporated genomic covariates through a genetics-conditioned prior adaptation. The hybrid decoder recovered treatment-effect parameters broadly consistent with previously reported nonlinear mixed-effects estimates, while achieving prior predictive performance comparable to the neural decoder. The joint model reproduced both tumor-volume distributions and dropout patterns in held-out individuals, and genetic conditioning improved individual-level prior predictions in both cutaneous melanoma and breast cancer experiments. Stability selection identified several biologically plausible genetic indicators, including alterations in BRAF, NRAS, NF1, and MDM2. These results demonstrate that EB-VAE provides a flexible probabilistic framework for combining neural dynamics, mechanistic structure, time-to-event modeling, and high-dimensional covariates in pharmacometric applications.
\end{abstract}
\section{Introduction}

Longitudinal measurements and time-to-event outcomes constitute two of the most important data modalities in pharmacometrics and quantitative systems pharmacology. Longitudinal biomarkers provide insight into disease progression and treatment response over time, whereas event outcomes such as dropout, progression, or survival capture clinically relevant endpoints that are often strongly associated with the underlying disease trajectory. Joint modeling approaches aim to integrate these complementary sources of information within a unified statistical framework, enabling improved inference and prediction while accounting for informative event processes \cite{mbogning2015joint,lindauer2017pembrolizumab,claret2009survival}.

At the same time, modern biomedical studies increasingly collect data from heterogeneous sources of information beyond longitudinal observations alone, including genomic profiling, molecular biomarkers, treatment histories, and other patient-specific covariates \cite{hasin2017multi,gao2015high}. These data sources provide additional context that may explain variability in treatment response and disease progression. Incorporating such multimodal information into longitudinal and time-to-event models remains a challenging problem, particularly when complex nonlinear relationships are present.

Traditional nonlinear mixed-effects (NLME) models provide a well-established framework for analyzing longitudinal and time-to-event data, allowing population-level trends to be separated from individual-specific variability \cite{lindstrom1990nlme,ribba2014review}. However, extending such models to incorporate high-dimensional covariates and flexible nonlinear representations can be challenging, often requiring substantial model development effort and computationally intensive inference procedures~\cite{ayral2021novel,sanghavi2024covariate}.

Machine learning methods have gained increasing attention in pharmacometrics and pharmacological modeling~\cite{lu2021neural,bram2025low,li2026variational,koch2020pharmacometrics,Rohleff2026Evolution}. In this context, neural ordinary differential equations (NODE) provide a natural way to combine neural network flexibility with continuous-time dynamical systems modeling~\cite{chen2018neural}, while variational autoencoders provide scalable amortized inference for latent-variable models~\cite{kingma2013auto,kingma2019vae}. However, integrating these ideas with covariate-conditioned population structure, time-to-event outcomes, and high-dimensional covariates (e.g., genomic information, images) remains an open methodological challenge.

In previous work, we introduced the empirical Bayes variational autoencoder (EB-VAE)~\cite{baaz2026ebvae}, a probabilistic latent-variable framework that shares several structural analogies with NLME modeling. The framework represents inter-individual variability (IIV) through latent individual effects regularized by a covariate-conditioned empirical Bayes prior, enabling scalable inference while retaining connections to classical population modeling concepts. Although EB-VAE models typically contain substantially more parameters than classical parametric population models, their effective complexity is controlled through variational and prior-based regularization, allowing them to remain flexible while mitigating overfitting. The EB-VAE framework is intentionally modular. Individual-specific latent effects are inferred through amortized variational inference, and the model components can incorporate a broad range of formulations, including fully neural representations, parametric models, or hybrid approaches. Additional outcome processes and heterogeneous data modalities can be integrated within the same probabilistic framework.

In this work, we further validate the EB-VAE framework by applying it to a substantially larger dataset, now consisting of patient-derived xenograft (PDX) tumor growth data. We further extend the framework to jointly model longitudinal tumor growth and time-to-event outcomes. Specifically, we incorporate a hazard-based time-to-event model to jointly describe tumor growth trajectories and time-to-dropout, investigate alternative decoder formulations including both fully neural and hybrid parametric–neural dynamics, and integrate genomic information through a dedicated feature extraction module. For the hybrid model, we obtain interpretable tumor dynamics parameters that are compared with values reported in the literature~\cite{baaz2022optimized}.


\section{Methods}

\subsection{Data}
\label{sec:data}
The analyses in this work are based on a large-scale PDX dataset comprising longitudinal tumor growth measurements, treatment information, and genomic data from 135 treatment regimens spanning multiple solid tumor types~\cite{gao2015high}. We focus our analyses on cutaneous melanoma (CM) and breast cancer (BC), consisting of 1342 tumor growth trajectories across 39 treatment regimens. The genomic data are available for most PDX models and include gene-level mutations, copy-number alterations, and RNA expression profiles.

We constructed a set of task-specific analysis datasets, summarized in Table~\ref{tab:analysis_datasets}, to evaluate different aspects of the proposed framework. The \emph{60-day 6-treatment CM} dataset comprises all CM PDXs from the untreated control group, as well as those treated with binimetinib, LEE001, encorafenib, or LEE011 in combination with either encorafenib or binimetinib. Tumor trajectories were truncated at day 60 to match the observation window used in a closely related NLME study~\cite{baaz2022optimized}, which modeled this subset of the data using an exponential tumor growth model with an additive drug effect. This dataset therefore provides a direct reference point for comparison with the proposed framework and the previously established NLME approach.

For the remaining datasets, only individuals with genetic data are kept, and the cutoffs are chosen to separate short-horizon (21 days) and longer-horizon (100 days) prediction settings. For all datasets, measurements after the specified cutoff are excluded. Individuals with measurements beyond the cutoff are treated as administratively censored at the cutoff time.






\begin{table}[htbp]
\centering
\small
\caption{Datasets used in the experimental evaluation. Cutoff refers to the last day for which observations were included. \(N\) denotes the number of individuals included in the dataset, Tx denotes the number of treatment groups, and Obs. denotes the average number of observations per individual, with the minimum and maximum numbers of observations given in brackets. The \emph{60-day 6-treatment CM} dataset contained the untreated, binimetinib, encorafenib, LEE011, LEE011 plus encorafenib, and LEE011 plus binimetinib treatment groups, whereas the remaining datasets contained all treatment groups available for the corresponding tumor type.}
\label{tab:analysis_datasets}
\begin{tabular*}{\textwidth}{@{\extracolsep{\fill}}lcccccc@{}}
\hline
\textbf{Dataset} & \textbf{Tumor}  & \textbf{Cutoff} & \textbf{\(N\)} & \textbf{Tx} & \textbf{Obs.} \\
\hline
\emph{60-day 6-treatment CM}
& CM
& 60 days
& 183
& 6
& 11.0 [3, 19] \\

\emph{21-day CM}
& CM
& 21 days
& 516
& 17
& 6.2 [3, 9] \\

\emph{100-day CM}
& CM
& 100 days
& 516
& 17
& 11.6 [3, 29] \\

\emph{21-day BC}
& BC
& 21 days
& 826
& 22
& 6.4 [2, 10] \\

\emph{100-day BC}
& BC
& 100 days
& 826
& 22
& 14.4 [2, 30] \\
\hline
\end{tabular*}
\end{table}

\subsection{Empirical Bayes VAE Framework}

To model the data, we use the EB-VAE framework~\cite{baaz2026ebvae}, which employs an encoder–decoder architecture to model both the dynamics and IIV in observed data. IIV is represented through latent variables $k_i$, which play a role analogous to individual-level parameters in traditional NLME models. Given longitudinal observations $y_i$ (the full observed time series for individual $i$), and individual covariates $x_i$ the encoder network approximates the posterior distribution
\begin{align*}
q_{\phi}(k_i \mid y_i, x_i),
\end{align*}
enabling inference of individual-specific latent effects. The resulting amortized inference procedure provides a scalable alternative to classical NLME estimation methods such as stochastic approximation expectation-maximization (SAEM)~\cite{delyon1999saem}. Once trained, the encoder maps individual observations and covariates directly to an approximate posterior distribution over latent effects.

The latent variables are regularized through a covariate-conditioned prior distribution,
\begin{align*}
p_{\psi}(k_i \mid x_i).
\end{align*}
The latent representation consists of covariate-explained structure and residual unexplained variation. The prior accounts for the former via a covariate-conditioned distribution, while the encoder captures the remaining variability.

The inferred latent variables are provided as individual-specific inputs to the decoder, which maps these latent representations to longitudinal tumor-volume trajectories. In the EB-VAE framework, the decoder defines the temporal evolution of a latent state $z_i(t)$ and its mapping to the longitudinal observations,
\begin{align*}
z_i(0)
&=
g_{\theta}(z_0,k_i),
\\
\dot z_i(t)
&=
f_{\theta}(z_i(t),k_i,u_i(t)),
\\
\hat y_i(t)
&=
h_{\theta}(z_i(t)).
\end{align*}
Here, $u_i(t)$ denotes treatment information when provided directly to the decoder, and $\hat y_i(t)$ denotes the predicted longitudinal measurement, here corresponding to tumor volume. The functions $g_{\theta}$, $f_{\theta}$, and $h_{\theta}$ may represent neural networks, mechanistic models, or hybrid formulations, with parameters $\theta$ shared across individuals. Depending on the decoder formulation, $k_i$ may either modulate a neural dynamical system or correspond directly to individual-level parameters in a mechanistic tumor-growth model.

The model is trained by maximizing the evidence lower bound (ELBO)~\cite{bishop2006prml}, jointly learning the encoder, decoder, and prior components within a unified probabilistic framework. A detailed description of the framework and training procedure is provided in \cite{baaz2026ebvae}.
\subsection{Joint Modeling}

We extend the framework to enable joint modeling of longitudinal (tumor volume) trajectories and the time-to-event processes (dropout). Let $C_{\max}$ denote the administrative censoring time and let $t_{i,\mathrm{last}}$ be the last available tumor measurement time for individual $i$. The observed follow-up time and event indicator are defined as
\begin{align*}
\tau_i
&=
\min(t_{i,\mathrm{last}}, C_{\max}),
\\
\delta_i
&=
\mathbf{1}\{t_{i,\mathrm{last}} < C_{\max}\}.
\end{align*}
Here, $\delta_i=1$ indicates dropout before the end of the analysis window, while $\delta_i=0$ indicates administrative right censoring. Given $k_i$ and $u_i$, the longitudinal and dropout processes are assumed conditionally independent,
\begin{align*}
p_{\theta}(y_i,\tau_i,\delta_i \mid k_i,u_i)
&=
p_{\theta}(y_i \mid k_i,u_i)
\,p_{\theta}(\tau_i,\delta_i \mid k_i,u_i),
\end{align*}
but remain coupled through the shared latent representation and decoder state.

\begin{figure}[htbp]
    \centering
\includegraphics[width=1\textwidth]{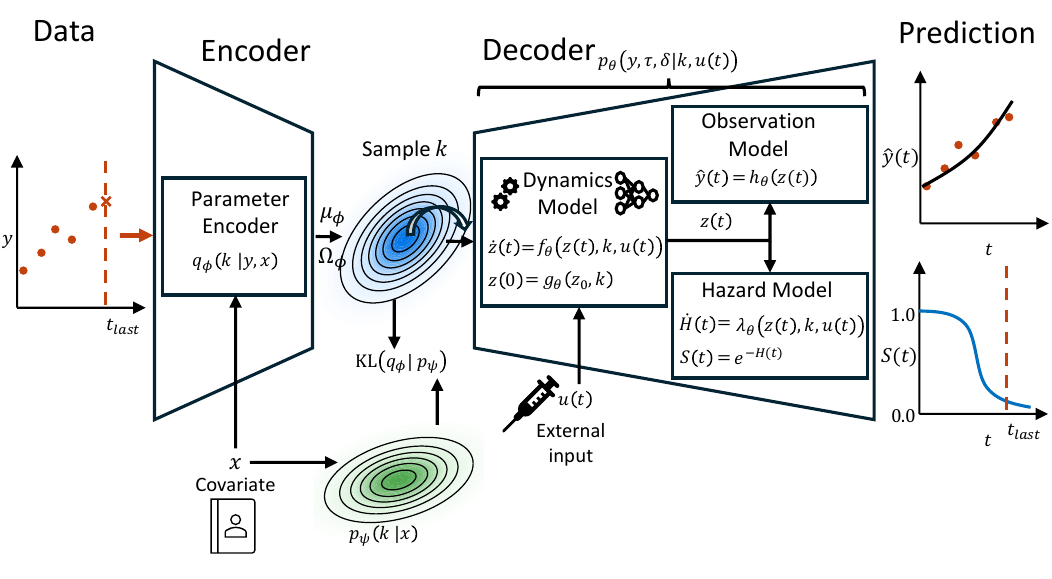}
    \caption{Overview of the empirical Bayes variational autoencoder framework for joint longitudinal and time-to-event modeling. Longitudinal tumor observations, treatment information, and optional covariates are used by the encoder to infer individual-specific latent effects $k_i$. A covariate-conditioned empirical Bayes prior $p_\psi(k_i \mid x_i)$ captures systematic population-level variability, including differences between treatment groups. Samples from the latent distribution are passed to a decoder that defines the longitudinal dynamics and predicts tumor volume trajectories. The decoder may be implemented either as a fully neural ODE model or as a mechanistic dynamical model. For joint modeling, the decoder is augmented with a hazard component that defines the cumulative event risk and survival probability over time.}
    \label{fig:ebjointvae}
\end{figure}

The event process is modeled through a time-dependent hazard function $\lambda_{\theta}(z_i(t),u_i(t),k_i)$. The corresponding cumulative hazard is defined as
\begin{align*}
H_i(t)
&=
\int_0^t
\lambda_{\theta}(z_i(s),u_i(s),k_i)
\,ds.
\end{align*}
This allows the cumulative hazard to be introduced as an additional state variable within the decoder dynamics. The augmented system, therefore, becomes
\begin{align*}
\frac{d}{dt}
\begin{bmatrix}
z_i(t)\\
H_i(t)
\end{bmatrix}
&=
\begin{bmatrix}
f_{\theta}(z_i(t),k_i,u_i(t))\\
\lambda_{\theta}(z_i(t),k_i,u_i(t))
\end{bmatrix}.
\end{align*}
The probability of remaining under observation up to time $t$ is given by
\begin{align*}
S_i(t)
&=
\exp\left(-H_i(t)\right).
\end{align*}
The resulting log-likelihood contribution for the time-to-event process is
\begin{align*}
\log p_{\theta}(\tau_i,\delta_i \mid k_i,u_i)
&=
-H_i(\tau_i)
+
\delta_i
\log
\lambda_{\theta}(z_i(\tau_i),k_i,u_i(\tau_i)).
\end{align*}
This likelihood term is incorporated directly into the EB-VAE objective, yielding a joint probabilistic model over both tumor trajectories and the time-to-event endpoint. Consequently, the latent variables are encouraged to capture factors associated with both disease progression and the duration of longitudinal follow-up. An overview of our framework can be seen in Figure \ref{fig:ebjointvae}.

\subsection{Decoder Formulations and Treatment-Effect Interpretation}
\label{sec:decoder-formulations}

The first formulation (Neural-EB-VAE) uses a fully neural decoder based on NODEs. In this setting, the functions $g_\theta$, $f_\theta$, and $h_\theta$ are parameterized by neural networks, providing a flexible data-driven representation.

The second formulation (Hybrid-EB-VAE) uses a hybrid decoder where the tumor dynamics are based on exponential tumor growth, whilst the hazard function is a NODE. In this formulation, the latent state corresponds directly to tumor volume and the observation model reduces to the identity mapping. The tumor dynamics are then given by, 
\begin{align*}
\frac{dz_i(t)}{dt}
&=
k_i z_i(t).
\end{align*}
$k_i$ in this formulation corresponds to an individual treatment-conditioned effective growth rate. 
Consequently, the population distribution of the latent growth parameter may vary between vehicle, monotherapy, and combination-treatment groups.

To enable comparison with classical tumor growth models, we translate the treatment-conditioned prior means into treatment-effect parameters. Let $\mu_a$ denote the prior mean of $k_i$ for treatment arm $a$. The vehicle group defines the baseline growth-rate parameter as
\begin{align*}
k_g
&=
\mu_{\mathrm{veh}}.
\end{align*}
For each treatment arm, the corresponding prior mean is parameterized as
\begin{align}
\mu_a
&=
k_g
- \sum_{j=1}^{M} a_j I_j(a)
- \sum_{j=1}^{M}\sum_{l>j}^{M} a_{jl} I_j(a) I_l(a),
\label{eq:treatment-decomposition}
\end{align}
where $I_j(a)$ indicates whether treatment $j$, with $j=1,\ldots,M$, is present in treatment arm $a$. The parameters $a_j$ represent the monotherapy effect of each drug, while $a_{jl}$ captures deviations from additivity (synergistic or antagonistic) of drug $j$ and $l$. Since each treatment arm consists of at most two drugs, higher-order interaction terms are not considered.

Importantly, these treatment-effect parameters are not explicit decoder parameters during training. They are instead obtained post hoc by solving the linear system implied by Eq.~\ref{eq:treatment-decomposition} using the treatment-specific prior means. This provides a direct way to compare the hybrid decoder with classical nonlinear mixed-effects tumor growth models while retaining the EB-VAE formulation in which treatment effects are represented through the covariate-conditioned prior.

\subsection{Genetic Covariate Representation and Prior Adaptation}
\label{sec:genetic-covariate-representation}

This section describes how genetic covariates were represented and incorporated into the EB-VAE prior. The available genetic annotations were sparse and heterogeneous and could therefore not be used directly as model inputs. We mapped them to a fixed-dimensional binary representation for each PDX tumor model and used this representation to condition the prior, allowing genetic information to influence prior predictions without retraining the learned longitudinal decoder dynamics.

Protein-changing mutations, amplifications, and deletions were encoded as binary indicators. Protein-changing mutations were represented at multiple levels of specificity, including exact protein changes, gene-level mutation indicators, missense indicators, and truncating-mutation indicators, whereas amplifications and deletions were represented as copy-number alteration indicators. The indicators were not mutually exclusive; for example, a specific amino-acid substitution could activate both an exact protein-change indicator and broader gene-level indicators for the same gene.

To obtain a sufficiently broad set of genetic variables, we combined literature-based selection, ChatGPT-assisted candidate generation, and random sampling. This procedure resulted in 463 genetic indicators that varied across PDX tumor models. Candidate generation was informed by the COSMIC Cancer Gene Census~\cite{sondka2018cosmic}, alterations highlighted by Gao et al.~\cite{gao2015high}, and literature on genes and alterations related to tumor biology, treatment response, and resistance mechanisms. Random sampling was used to broaden the candidate set and evaluate whether the model could identify informative features beyond the biologically informed candidates.

The genetic input was high-dimensional, with many more candidate genetic indicators than distinct PDX tumor models. Direct end-to-end training with genetic covariates could therefore lead to unstable learning and overfitting. We therefore incorporated genetic information using a two-stage training procedure. In the first stage, the longitudinal encoder--decoder model and the treatment-conditioned empirical Bayes prior were trained, while the prior was not conditioned on genetic covariates. This step established the latent individual-effect space. In the second stage, the encoder, decoder, and treatment-conditioned prior were kept fixed, and only a neural shift model conditioned on genetic covariates and treatment was trained. The shift model modified the empirical Bayes prior in the learned latent space, thereby defining a genetics-conditioned empirical Bayes prior and allowing genetic effects to depend on treatment context. The shift model was optimized through the Kullback--Leibler divergence (KL) term between the fixed encoder and the genetics-conditioned empirical Bayes prior, without changing the learned decoder dynamics.

Stability selection was used as an exploratory tool for ranking genetic indicators associated with the genetics-conditioned prior~\cite{meinshausen2010stability}. In this framework, genetic indicators can be ranked by perturbing individual indicators and measuring the resulting change in the validation KL divergence between the encoder and the genetics-conditioned prior. Repeating this procedure over perturbed datasets provides a stability score for each indicator, interpreted as a model-based and hypothesis-generating measure of how consistently the indicator influences the learned prior distribution.

\subsection{Evaluation Metrics and Visual Diagnostics}
Visual predictive checks (VPCs), Kaplan--Meier VPCs (KM--VPCs), and prediction corrected VPCs (pcVPCs)~\cite{bergstrand2011pcvpc} were used to qualitatively assess whether prior predictive simulations reproduced the observed tumor volume and dropout distributions over time. Moreover, individual-level root mean squared error (RMSE) between observed tumor volumes and the median prior/posterior predictive trajectory of held-out individuals was used to quantify predictive performance. RMSE was computed per individual and then aggregated across individuals, giving each individual equal weight. To assess prediction performance over different follow-up horizons, cumulative RMSE curves were computed by progressively including later observation times. Detailed definitions of the RMSE metrics and visual diagnostic procedures are provided in the Supporting Information.

\subsection{Experiments}
\label{sec:experiments}

We conduct four experiments to evaluate complementary aspects of the proposed framework. 

\experiment{Experiment 1}{Mechanistic interpretation and decoder comparison}
The first experiment evaluates both the interpretability of the Hybrid-EB-VAE and its predictive performance relative to the Neural-EB-VAE using the \emph{60-day 6-treatment CM} dataset. First, the Hybrid-EB-VAE is trained five times with the full dataset and the learned treatment-specific prior means are mapped onto the treatment-effect decomposition in Eq.~\ref{eq:treatment-decomposition}. The resulting parameters are compared across runs and the means are compared with literature values. Model adequacy is assessed using treatment-stratified VPCs and KM VPCs. Secondly, the predictive performance of both models is evaluated using five-fold cross-validation with the \emph{60-day 6-treatment CM} dataset. 

\experiment{Experiment 2}{Population prediction and treatment generalization}

The second experiment evaluates the ability of the Neural-EB-VAE to generalize to unseen individuals and treatment groups in the \emph{100-day CM} and \emph{100-day BC} datasets. 
Individual-holdout prediction was evaluated using five-fold cross-validation. Predictive performance was assessed using pcVPCs constructed by pooling held-out individuals across treatment groups, together with pooled KM VPCs for the dropout process. As a reference, we also evaluated a naïve dropout rule in which dropout occurred when the simulated tumor volume reached 1500~mm$^3$.
In addition, treatment-holdout prediction was performed by excluding one treatment group during training and using it only for evaluation. The held-out treatment group was either the untreated group or a treatment selected such that the drug was represented elsewhere in the training data, either as a monotherapy or as part of a combination therapy. Performance on excluded treatments was assessed using treatment-specific VPCs.

\experiment{Experiment 3}{Genetic covariates and feature stability analysis}

The third experiment investigated whether genetic covariates improved predictive performance and which genetic indicators contributed most consistently to the genetics-aware prior. Models were compared before and after including the genetic covariates. The comparison was performed on both \emph{21-day} and \emph{100-day} datasets. Individual-level RMSE was used to quantify overall predictive accuracy, while cumulative individual RMSE was used to evaluate how the benefit of genetic conditioning changed as longer follow-up horizons were included. Stability selection was used to characterize the robustness of the identified genetic indicators.

\experiment{Experiment 4}{Benefit of multi-treatment training}
The fourth experiment assessed whether jointly modeling multiple treatment groups improved prediction for individual treatment groups. This experiment used the \emph{21-day} datasets. For each selected treatment group, a treatment-specific model was trained using only individuals from that group, while another model was trained using the same treatment-specific individuals together with individuals from the remaining treatment groups. Both models were evaluated on the same held-out individuals from the selected treatment group to directly compare the benefit of multi-treatment training.

\section{Results}

\subsection{Mechanistic interpretation and decoder comparison}

For the Hybrid-EB-VAE, the baseline (untreated) growth rate was estimated with limited variation across the five runs \([0.0820,0.0844]\). The mean value was \(k_g = 0.0830\), which is slightly higher than the reference value (\(k_g = 0.06\)) reported by \cite{baaz2022optimized}. The estimated monotherapy parameter for LEE011, encorafenib, and binimetinib were \(0.0372\), \(0.0340\), and \(0.0212\), respectively, compared with reference values of \(0.0444\), \(0.0393\), and \(0.0090\). Overall, the estimates are fairly close to the reference values. However, it should also be noted that the reference model did not account for the dropout process and explicitly modeled the impact of mutations in the BRAF-gene, which may contribute to differences in the estimated growth dynamics. Furthermore, the learned combination-deviation terms were small, indicating limited interaction effects, consistent with the reference model.

\begin{figure}[!hbtp]
    \centering
    \includegraphics[width=0.98\textwidth]{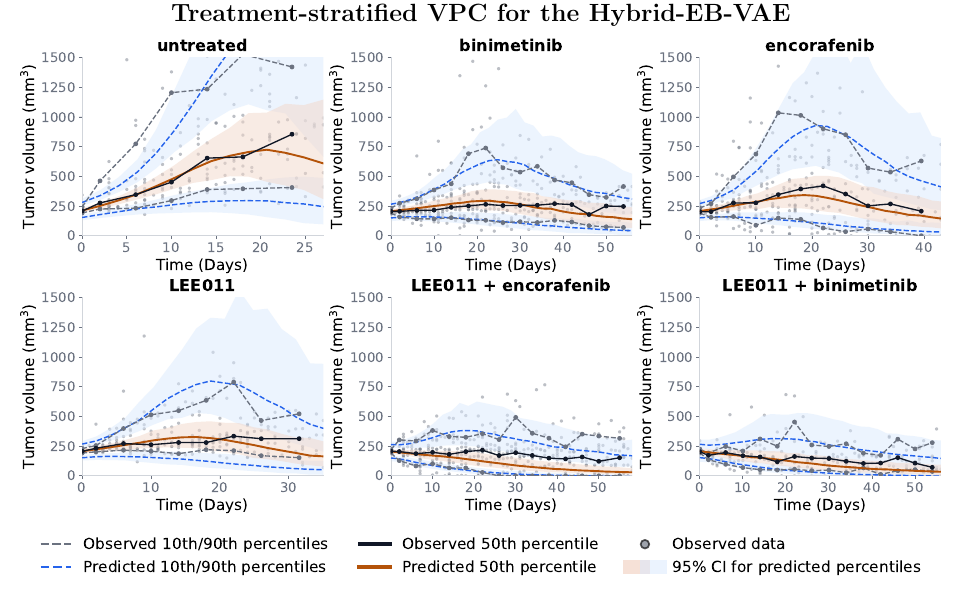}

    \caption{Treatment-stratified VPCs for the Hybrid-EB-VAE, illustrating model fit to the observed tumor-volume data. Observed percentiles are shown in black/gray and predicted percentiles in blue/orange, with shaded regions indicating 95\% confidence intervals for the predicted percentiles. Simulated trajectories were censored according to the learned dropout process, and VPCs were truncated once fewer than 10 individuals remained under observation within a treatment group to avoid unstable empirical percentile estimates at late time points.}
    \label{fig:hybrid-full-training-vpc}
\end{figure}

The treatment-stratified VPCs in Figure~\ref{fig:hybrid-full-training-vpc} show that the hybrid model reproduced the main tumor-volume patterns across treatment groups. Supplementary KM-VPCs showed that the learned dropout component also captured the treatment-specific probability of remaining under observation over time.

In the comparison between the Hybrid-EB-VAE and the Neural-EB-VAE, the two models showed nearly identical performance for prior predictions on held-out individuals, with mean RMSE values of 195 for the Hybrid-EB-VAE and 196 for the Neural-EB-VAE. For posterior predictions, the Neural-EB-VAE achieved the lowest mean RMSE (68 versus 97). These results suggest that the mean tumor growth dynamics are well approximated by a simple exponential growth model, despite the more complex and heterogeneous trajectories observed at the individual level. We therefore use the Neural-EB-VAE in the remaining population-level prediction experiments, as it provides greater modeling flexibility and does not require separate mechanistic assumptions for different treatment settings or treatment combinations.

\subsection{Population prediction and treatment generalization}

Figure~\ref{fig:fold0-dropout-vpc-summary} compares the population-level prior predictive performance of the Neural-EB-VAE and the naïve threshold-based dropout model for held-out individuals in one representative cross-validation split of the \emph{100-day CM} dataset. The figure shows KM-VPCs for dropout and pcVPCs for tumor volume. The remaining cross-validation splits exhibited similar behavior and are provided in the supplementary material, together with the corresponding results for the \emph{100-day BR} dataset. The Neural-EB-VAE accurately captured the observed dropout pattern, and its pcVPC showed good agreement between the observed and simulated tumor-volume distributions. In contrast, the naïve threshold-based dropout model produced biased dropout predictions, resulting in noticeable discrepancies in the tumor-volume pcVPC.

\begin{figure}[!htbp]
  \centering

  \includegraphics[width=1\textwidth]{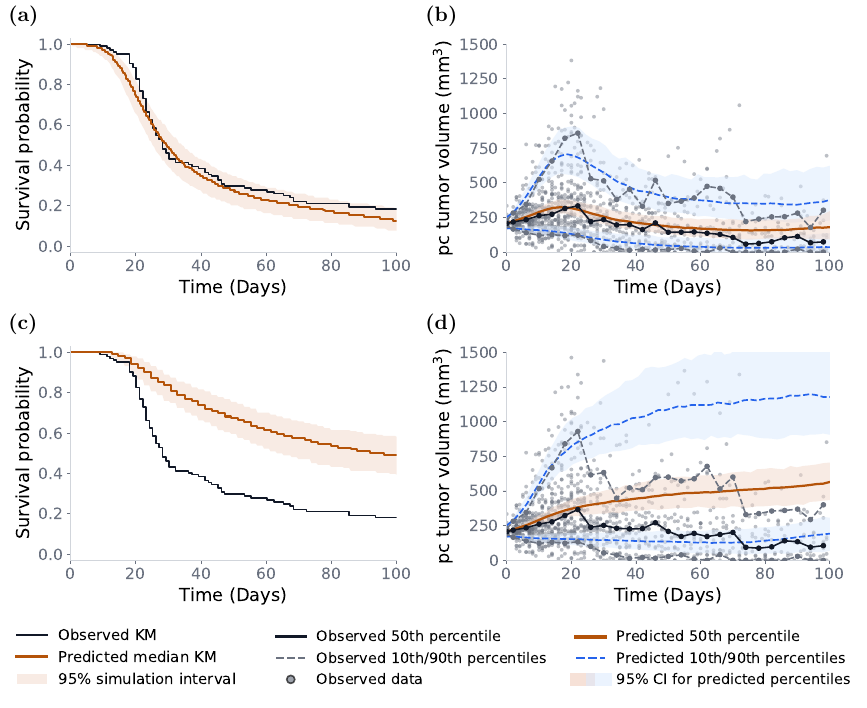}

  \caption{Kaplan–Meier visual predictive checks (KM-VPCs; panels a,c) and prediction-corrected visual predictive checks (pcVPCs; panels b,d) for a representative cross-validation split of the cutaneous melanoma dataset, with test individuals pooled across treatment groups. Observed percentiles (not used for training) are shown in black and gray, while predicted percentiles are shown in blue and orange. Shaded regions indicate 95\% confidence intervals for the predicted percentiles.
(a) KM-VPC using the learned survival-model dropout mechanism.
(b) pcVPC using the learned survival-model dropout mechanism.
(c) KM-VPC using the naïve threshold-based dropout mechanism.
(d) pcVPC using the naïve threshold-based dropout mechanism.
}
  \label{fig:fold0-dropout-vpc-summary}
\end{figure}

\begin{figure}[!htbp]
  \centering

  \includegraphics[width=1\textwidth]{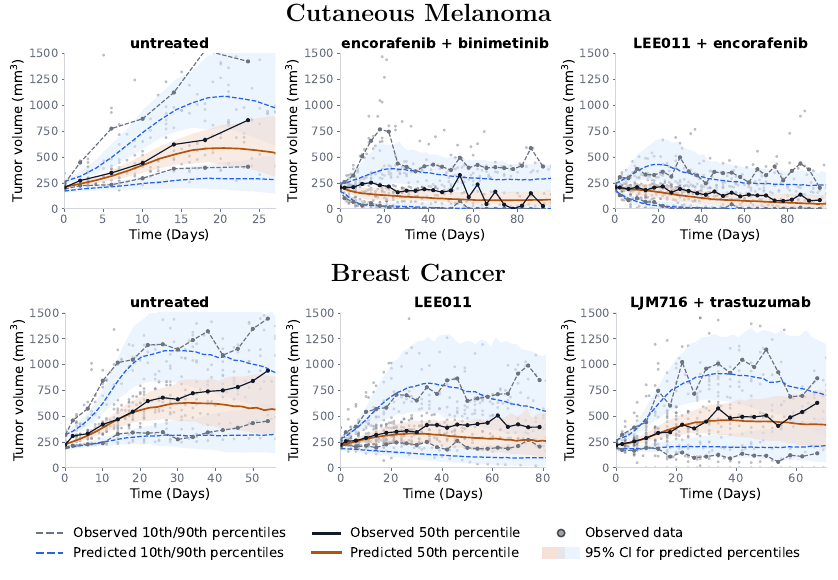}

    \caption{Treatment-holdout VPCs for the cutaneous melanoma (first row) and breast cancer (second row) datasets. Each panel evaluates prior predictive performance for a treatment group excluded during training. For each cancer type, panels were selected to show the untreated group, the poorest-fitting held-out treatment group, and the best-fitting held-out treatment group among the evaluated treatment-holdout experiments.}
  \label{fig:vpc-treatment-holdout-cm-bc}
\end{figure}

Figure~\ref{fig:vpc-treatment-holdout-cm-bc} shows the population-level prediction performance for held-out treatment groups using the Neural-EB-VAE. The model was able to reproduce the main longitudinal patterns for several unseen treatments and more examples are provided in the supplementary information.

\subsection{Genetic covariates and feature stability analysis}

\begin{figure}[htbp]
    \centering

    \includegraphics[width=1\textwidth]{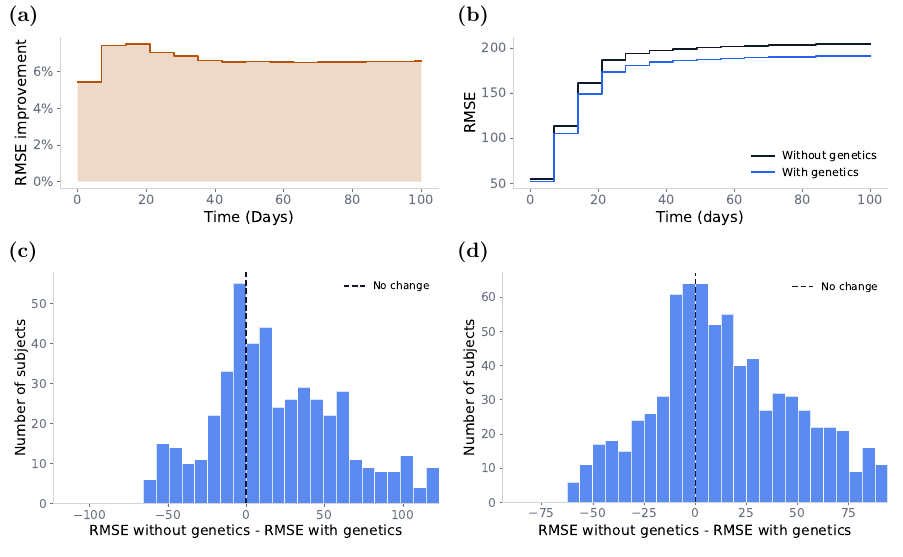}

    \caption{Comparison of prior predictive performance with and without genetic conditioning.
(a) Cumulative mean individual RMSE improvement for the 100-day CM experiment, expressed as percent improvement from conditioning the prior on genetic information.
(b) Cumulative mean individual RMSE with and without genetic conditioning for the 100-day CM experiment.
(c) Distribution of individual RMSE improvements for the experiment based on the \emph{21-day CM} dataset.
(d) Distribution of individual RMSE improvements for the experiment based on the \emph{21-day BC} dataset.
The histograms in panels (c) and (d) had the 5\% smallest and largest values removed before plotting.
Individual RMSE values were averaged over five runs with genetics and five runs without genetics.}
    \label{fig:genetics-rmse-summary}
\end{figure}

We next evaluated whether conditioning the model on genetic covariates improved prior prediction for held-out individuals. Figure~\ref{fig:genetics-rmse-summary} shows the cumulative individual RMSE for the 100-day CM experiment. Genetic conditioning reduced RMSE throughout the follow-up period, with the largest differences occurring early and remaining relatively stable thereafter. 

For the \emph{21-day} datasets genetic conditioning resulted in improved individual-level predictions (Fig~\ref{fig:genetics-rmse-summary}). Across five runs with genetics and five runs without genetics, mean individual RMSE decreased from 161 to 140 (13\% reduction) and of the 516 held-out individuals, 325 showed improved predictions when genetic covariates were included. The results for the \emph{21-day BC} dataset showed a similar trend (Fig~\ref{fig:genetics-rmse-summary}). Full results are provided in the Supplementary Information.

The top-ranked indicators from the stability selection included BRAF V600E, which is a known melanoma-associated alteration~\cite{bharti2025braf}. Several other highly selected indicators, including NRAS, NF1, and MDM2 missense indicators, were also consistent with genes involved in growth signaling, cell-cycle regulation, or growth inhibition~\cite{arnoff2022mdm2,guo2021signal}. A complete list of the top 10 ranked indicators is provided in the Supplementary Information.

\subsection{Benefit of multi-treatment training}

\begin{figure}[!htbp]
    \centering
    \includegraphics[width=1\textwidth]{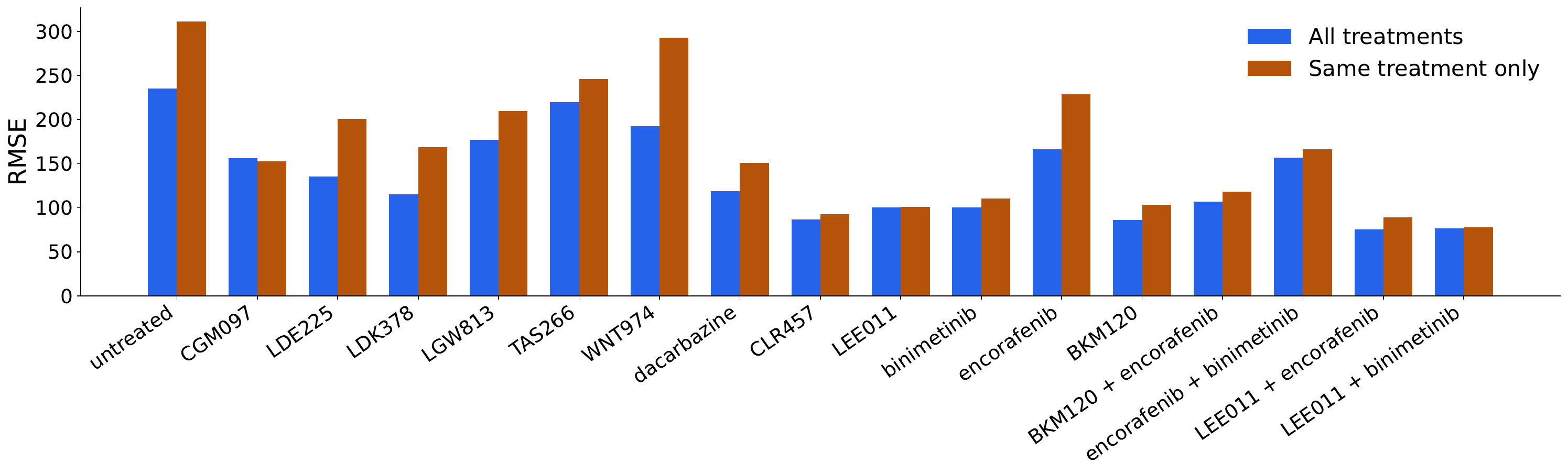}
    \caption{Mean individual prior RMSE for selected cutaneous melanoma treatment groups under two training regimes. 
    ``Same treatment only'' denotes models trained only on individuals from the evaluated treatment group. 
    ``All treatments'' denotes models trained on the same treatment-specific training individuals together with individuals from the remaining treatment groups. 
    Both models were evaluated on the same held-out individuals from the treatment group shown on the x-axis.}
    \label{fig:data-gen-cm}
\end{figure}

Figure~\ref{fig:data-gen-cm} shows the effect of jointly training across treatment groups on prior predictive performance for individual CM treatment groups. For each treatment group, we compared a treatment-specific model trained exclusively on individuals from that group with a jointly trained model that additionally incorporated individuals from the remaining treatment groups. Both models were evaluated on the same held-out individuals from the corresponding treatment group. Joint training improved prediction across almost all treatment groups. The largest improvements were observed for the untreated group, dacarbazine, and encorafenib, whereas the LEE011--binimetinib group showed only a minor improvement. CGM097 was the only group that did not benfit from joint training. 

\section{Discussion}
In this work, we evaluated the EB-VAE framework across a broader range of oncology modeling tasks, demonstrating its flexibility for joint longitudinal and time-to-event modeling, genetic covariate integration, and alternative decoder formulations. A key extension was the incorporation of a joint longitudinal and time-to-event decoder that links tumor growth and dropout through a shared latent representation and NODE-based hazard function. In the Neural-EB-VAE, tumor growth was modeled using a NODE, whereas the Hybrid-EB-VAE combined an interpretable exponential tumor-growth model with the same NODE-based hazard formulation. Trained on the \emph{60-day 6-treatment CM} data the Hybrid-EB-VAE recovered tumor-growth parameters broadly consistent with previously reported NLME estimates~\cite{baaz2022optimized}, while the Neural-EB-VAE achieved improved overall predictive performance for the cross-validation. Together, these results demonstrate that the EB-VAE framework provides a modular approach for integrating mechanistic and data-driven components within a unified latent-variable model.

The incorporation of genomic information further demonstrated the ability of the EB-VAE framework to leverage high-dimensional patient-specific data. In the present study, the genetic feature model was evaluated together with the Neural-EB-VAE, where it improved predictive performance on held-out individuals by up to 13\% in terms of RMSE. Examination of individual prediction errors indicated that these improvements were observed across the majority of individuals. Most of the improvement was observed at the earlier time points, indicating that the model was able to effectively leverage the larger amount of available longitudinal information to learn how genomic features influence tumor growth dynamics. At later time points, the benefits of genomic information were reduced, likely due to increasing data sparsity resulting from dropout, which limits the amount of observed information available for learning and prediction.

Moreover, the genetic representation used in this work was limited to a subset of the available genomic information, and incorporating additional genetic indicators would require further feature processing and model development. Therefore, these results should be viewed as a proof-of-concept demonstrating the potential of integrating genomic covariates within the EB-VAE framework. As larger and more comprehensive genomic datasets become available, enabling the inclusion of additional relevant markers, further improvements in predictive performance may be expected. One could also imagine using a pre-trained network to generate more informative genomic embeddings, similar to approaches used in other high-dimensional biological data domains, where learned representations capture complex patterns beyond manually selected features. 

The two-stage genetic adaptation used here was introduced to stabilize learning in a setting with many candidate genetic indicators and comparatively few distinct PDX tumor models. With substantially larger datasets, the empirical Bayes prior could instead be conditioned jointly on treatment and genetic covariates during end-to-end training. Beyond improving predictive performance, the proposed stability-selection procedure provides a means of identifying genetic features that are consistently utilized by the model across repeated training runs. In this sense, the genomic feature model serves not only as a mechanism for incorporating high-dimensional molecular data, but also as a data-driven covariate selection approach that may help prioritize candidate biomarkers for further biological investigation.

To further assess generalization performance, we considered a leave-one-treatment-arm-out cross-validation setup, in which the Neural-EB-VAE was trained on all but one treatment regimen and evaluated on the held-out arm. This setting requires extrapolation to an unseen therapeutic condition rather than interpolation within the observed treatment distribution. The results show that the framework is able to produce reasonable predictions for previously unseen treatment arms, indicating that the learned treatment representations capture structured effects that can transfer beyond the observed combinations. This suggests that, with sufficiently diverse training data covering a broad range of monotherapies and combination therapies, the framework may be capable of supporting extrapolation to new treatment regimens.

While the proposed framework demonstrates strong performance across a range of experimental settings, its current evaluation is primarily based on two of the six available tumor types, and therefore represents a proof-of-concept rather than a fully comprehensive model trained across all available data. The experiments in which we jointly modeled several treatment arms yielded promising results, suggesting that pooling data across related experimental conditions can improve the learned representations. This idea could naturally be extended to jointly model several cancer types, for example through a shared or hierarchical structure, which is expected to further improve the robustness of the learned representations and may lead to enhanced predictive performance across tumor types. In this broader setting, the genomic feature extraction component will likely require further development to ensure that it can generalize effectively when learning from multiple tumor types simultaneously.

\section*{Study Highlights}

\begin{itemize}

\item[\textbf{What is the current knowledge on the topic?}]

Population models are widely used to describe tumor growth and treatment response, but conventional nonlinear mixed-effects approaches can be difficult to extend to complex longitudinal dynamics, informative dropout, and high-dimensional covariates such as genomic data. Machine learning methods offer flexible alternatives, but their integration with pharmacometric population modeling remains challenging.

\item[\textbf{What question did this study address?}]

We asked whether the empirical Bayes variational autoencoder (EB-VAE) framework could be extended to jointly model longitudinal tumor growth, dropout, alternative decoder formulations, and genomic covariates in patient-derived xenograft studies.

\item[\textbf{What does this study add to our knowledge?}]

The study shows that EB-VAE can combine neural and hybrid tumor-growth models within the same population modeling framework, jointly describe tumor growth and dropout, and use genomic covariates to improve individual-level prior predictions. The framework also enabled exploratory identification of genetic indicators associated with the learned prior adaptation.

\item[\textbf{How might this change drug discovery, development, and/or therapeutics?}]

This approach may support more flexible analysis of preclinical oncology studies by integrating tumor-growth dynamics, dropout, treatment information, and molecular covariates in a single probabilistic framework. Such models could help prioritize treatment-response hypotheses and guide future model-informed drug discovery and development.

\end{itemize}

\section*{Acknowledgments}

ChatGPT was used to improve wording and readability; the authors have reviewed all content and take full responsibility for the final text.

Cancer gene annotations were obtained from the COSMIC Cancer Gene Census (CGC), COSMIC v103, accessed 12 May 2026. The downloaded CGC table was used to annotate genes according to their curated cancer relevance.

\section*{Funding}
No funding was received for this work.

\section*{Conflicts of Interest}
The authors declare no conflicts of interest.

\section*{Author Contributions}
AS, NO, MB, and MJ wrote the manuscript;  
AS, NO, and MB designed the research; 
NO performed the research; 
AS, NO, and MB analyzed the data.

\bibliographystyle{ieeetr}
\bibliography{references}
\section*{Appendix}
\appendix

\section{Detailed Evaluation Procedures}
\label{sec:supp-evaluation}
Models were evaluated using both prior and posterior predictive simulations, depending on the objective of the analysis. Prior predictive simulations were used to assess population-level predictive performance for held-out individuals based only on covariates and treatment information. Posterior predictive simulations were used to assess the ability of each model to individualize tumor-growth trajectories after conditioning on the observed longitudinal data for a given individual. Thus, prior predictions evaluate population-level generalization, whereas posterior predictions provide a reconstruction and individualization diagnostic.

RMSE-based metrics were used to quantify predictive performance, while visual diagnostics were used to assess whether simulated trajectories reproduced the observed tumor-volume and dropout patterns. Unless otherwise stated, evaluations were performed on held-out individuals from the corresponding experiment. For posterior predictive evaluations, the held-out individual was not used during model training, but the individual's observed data were used at evaluation time to infer the corresponding latent representation through the encoder.

For a given split $s$ of a dataset $\mathcal D$, let $i\in\mathcal D_{\mathrm{test}}^{(s)}$ denote a held-out test individual, and let $\mathcal T_i=\{t_{i1},\ldots,t_{in_i}\}$ denote the corresponding ordered observation times. For each predictive replicate $r=1,\ldots,R$, latent variables were sampled either from the covariate-conditioned prior or from the approximate posterior. In prior predictive evaluations, samples were drawn as
\begin{align*}
k_i^{(r)}
&\sim
p_{\psi}(k_i \mid x_i),
\end{align*}
whereas in posterior predictive evaluations, samples were drawn from the encoder distribution,
\begin{align*}
k_i^{(r)}
&\sim
q_{\phi}(k_i \mid y_i, x_i).
\end{align*}
The sampled latent variables were then propagated through the decoder to obtain a simulated tumor-volume trajectory $\tilde y_i^{(r)}(t)$.

For joint models with a time-to-event component, the dropout process was simulated from the learned hazard model in each predictive replicate. This produced a simulated dropout time for each individual and replicate. We let
\begin{align*}
R_i^{(r)}(t)
&\in
\{0,1\}
\end{align*}
indicate whether individual $i$ has not dropped out by time $t$ in predictive replicate $r$. Thus, $R_i^{(r)}(t)=1$ before the simulated dropout time and $R_i^{(r)}(t)=0$ after dropout. For models without a time-to-event component, $R_i^{(r)}(t)=1$ for all $t$.

The retention-aware pointwise median predictive trajectory was computed from the simulated trajectories that remained under observation at each time point,
\begin{align*}
\tilde y_{i,\mathrm{med}}(t)
&=
\operatorname{median}
\left\{
\tilde y_i^{(r)}(t)
:
R_i^{(r)}(t)=1
\right\}.
\end{align*}
This definition ensures that tumor-volume predictions are summarized for the simulated population still under observation, matching the observation process represented in the data.

The following subsections define the specific evaluation metrics and visual diagnostics used in the experiments. Some quantitative metrics, such as RMSE, naturally aggregate across cross-validation folds. Let $\mathcal S$ denote the set of splits used in a given experiment, and define the set of held-out individual--split pairs as
\begin{align*}
\mathcal I_{\mathrm{test}}
&=
\left\{
(s,i): s\in\mathcal S,\ i\in\mathcal D_{\mathrm{test}}^{(s)}
\right\}.
\end{align*}
In the cross-validation experiments, each individual appears in the test set exactly once, so aggregating over $\mathcal I_{\mathrm{test}}$ corresponds to aggregating over the individuals in the dataset.

\subsection*{Individual-level RMSE}

Individual-level RMSE was used to quantify out-of-sample prediction error for held-out individuals while giving each individual equal weight, rather than weighting individuals in proportion to their number of observations. To evaluate prediction performance over different follow-up horizons, we define the metric using a cutoff time $t_c$.

For individual $i$, let
\begin{align*}
\mathcal T_i(t_c)
&=
\left\{
t\in\mathcal T_i : t\leq t_c
\right\}
\end{align*}
denote the observed measurement times up to the cutoff, and let
\begin{align*}
n_i(t_c)
&=
|\mathcal T_i(t_c)|.
\end{align*}
For individuals with at least one observation up to $t_c$, the cutoff-specific individual RMSE for individual $i$ in split $s$ was defined as
\begin{align*}
\mathrm{RMSE}_i^{(s)}(t_c)
&=
\sqrt{
\frac{1}{n_i(t_c)}
\sum_{t\in\mathcal T_i(t_c)}
\left(
y_i(t)
-
\tilde y_{i,\mathrm{med}}^{(s)}(t)
\right)^2
}.
\end{align*}

To aggregate across cross-validation splits, we defined the cutoff-specific set of evaluable held-out individual--split pairs as
\begin{align*}
\mathcal I_{\mathrm{test}}(t_c)
&=
\left\{
(s,i)\in\mathcal I_{\mathrm{test}}
:
n_i(t_c)>0
\right\}.
\end{align*}
Overall performance at cutoff $t_c$ was summarized by the mean individual RMSE,
\begin{align*}
\mathrm{RMSE}(t_c)
&=
\frac{1}{|\mathcal I_{\mathrm{test}}(t_c)|}
\sum_{(s,i)\in\mathcal I_{\mathrm{test}}(t_c)}
\mathrm{RMSE}_i^{(s)}(t_c),
\end{align*}
and by the median individual RMSE,
\begin{align*}
\mathrm{median\ RMSE}(t_c)
&=
\mathrm{median}
\left\{
\mathrm{RMSE}_i^{(s)}(t_c)
:
(s,i)\in\mathcal I_{\mathrm{test}}(t_c)
\right\}.
\end{align*}

Evaluating these quantities over a sequence of increasing cutoff times gives a cumulative individual RMSE curve, showing how prediction error changes as progressively later observations are included.

\subsection*{Visual Predictive Checks}
\label{sec:vpc}

Visual predictive checks were used to assess whether the prior predictive distribution reproduced the observed distribution of tumor volumes over time. To avoid repeating the same construction for standard and prediction-corrected visual predictive checks, we define the diagnostic in terms of generic observed and simulated values. Let $v_i(t)$ denote the observed value used in the diagnostic and let $\tilde v_i^{(r)}(t)$ denote the corresponding simulated value in replicate $r$. For a standard visual predictive check,
\begin{align*}
v_i(t)
&=
y_i(t),
&
\tilde v_i^{(r)}(t)
&=
\tilde y_i^{(r)}(t).
\end{align*}
For prediction-corrected visual predictive checks, $v_i(t)$ and $\tilde v_i^{(r)}(t)$ instead denote the corresponding prediction-corrected observed and simulated values.

Let $\mathcal D_{\mathrm{eval}}$ denote the individuals included in the diagnostic. For treatment-stratified visual predictive checks, let $a$ index treatment groups and let $x_i^{\mathrm{treat}}$ denote the treatment group of individual $i$. The follow-up interval was divided into non-overlapping time bins
\begin{align*}
\mathcal B
&=
\{B_1,\ldots,B_K\}.
\end{align*}
For treatment group $a$ and time bin $B_b$, the observed values were collected as
\begin{align*}
\mathcal V_{a,b}^{\mathrm{obs}}
&=
\left\{
v_i(t):
i\in\mathcal D_{\mathrm{eval}},\,
x_i^{\mathrm{treat}}=a,\,
t\in\mathcal T_i,\,
t\in B_b
\right\}.
\end{align*}
Observed values are included only at recorded measurement times and therefore already reflect the observed dropout process.

For each prior predictive replicate $r$, the corresponding simulated values were collected as
\begin{align*}
\mathcal V_{a,b}^{(r)}
&=
\left\{
\tilde v_i^{(r)}(t):
i\in\mathcal D_{\mathrm{eval}},\,
x_i^{\mathrm{treat}}=a,\,
t\in B_b,\,
R_i^{(r)}(t)=1
\right\}.
\end{align*}
Thus, for joint models, simulated values were included only for individuals that had not dropped out in the corresponding predictive replicate.

For a percentile level $p$, the observed percentile in treatment group $a$ and bin $B_b$ was computed as
\begin{align*}
q_{p,a,b}^{\mathrm{obs}}
&=
Q_p\left(\mathcal V_{a,b}^{\mathrm{obs}}\right),
\end{align*}
where $Q_p(\cdot)$ denotes the empirical $p$-th percentile. The corresponding simulated percentile was computed separately for each replicate,
\begin{align*}
q_{p,a,b}^{(r)}
&=
Q_p\left(\mathcal V_{a,b}^{(r)}\right).
\end{align*}
The visual predictive check displays the observed percentiles together with the median simulated percentiles,
\begin{align*}
\tilde q_{p,a,b}
&=
\operatorname{median}
\left\{
q_{p,a,b}^{(r)}: r=1,\ldots,R
\right\},
\end{align*}
and simulation intervals obtained from the empirical quantiles of
\begin{align*}
\left\{
q_{p,a,b}^{(r)}
:
r=1,\ldots,R
\right\}.
\end{align*}
Bins with too few observed individuals were excluded from the diagnostic. In treatment-stratified visual predictive checks, panels were truncated once fewer than 10 observed individuals remained before dropout in the corresponding treatment group, to avoid interpreting empirical percentiles based on very small sample sizes at late time points.

In this work, standard uncorrected visual predictive checks were used as treatment-stratified training-data model-adequacy diagnostics. For out-of-sample diagnostics in which treatment groups were pooled, prediction-corrected visual predictive checks were used.

\subsection*{Kaplan--Meier Visual Predictive Checks for Dropout}
\label{sec:km-vpc}

Kaplan--Meier visual predictive checks were used to assess whether the simulated dropout process reproduced the observed probability of not having dropped out over time. For each individual $i$, let $\tau_i$ denote the observed follow-up time and let $\delta_i\in\{0,1\}$ denote the dropout indicator, where $\delta_i=1$ indicates observed dropout and $\delta_i=0$ indicates administrative censoring.

Let $\mathcal D_{\mathrm{eval}}$ denote the individuals included in the diagnostic. Depending on the experiment, $\mathcal D_{\mathrm{eval}}$ was either the fitted training dataset or the pooled held-out test individuals across cross-validation splits. For treatment-stratified diagnostics, let $a$ index treatment groups. The observed Kaplan--Meier curve for treatment group $a$ was estimated from
\begin{align*}
\left\{
(\tau_i,\delta_i):
i\in\mathcal D_{\mathrm{eval}},\,
x_i^{\mathrm{treat}}=a
\right\}.
\end{align*}

For each prior predictive replicate $r$, dropout times were simulated from the learned cumulative hazard. Specifically, an exponential threshold was sampled as
\begin{align*}
E_i^{(r)}
&\sim
\mathrm{Exp}(1),
\end{align*}
and the simulated dropout time was defined as the first time at which the cumulative hazard exceeded this threshold. The simulated dropout process was then converted into a simulated follow-up time $\tau_i^{(r)}$ and event indicator $\delta_i^{(r)}$. If simulated dropout occurred during the follow-up interval, $\tau_i^{(r)}$ was set to the simulated dropout time and $\delta_i^{(r)}=1$. If no simulated dropout occurred before the end of follow-up, the individual was treated as censored at the end of follow-up and $\delta_i^{(r)}=0$.

The simulated Kaplan--Meier curve for replicate $r$ and treatment group $a$ was estimated from
\begin{align*}
\left\{
(\tau_i^{(r)},\delta_i^{(r)}):
i\in\mathcal D_{\mathrm{eval}},\,
x_i^{\mathrm{treat}}=a
\right\}.
\end{align*}
Across simulation replicates, the diagnostic displays the observed Kaplan--Meier curve together with the median simulated curve and a simulation interval obtained from the empirical quantiles of the simulated Kaplan--Meier curves at each time point. This diagnostic was used to evaluate whether the learned time-to-event component reproduced the observed dropout dynamics.

\subsection*{Prediction-Corrected Visual Predictive Checks}
\label{sec:pcvpc}

Prediction-corrected visual predictive checks were used for population-level visual diagnostics when multiple treatment groups were pooled. Since different treatment groups can have different expected tumor-volume trajectories, direct pooling may obscure whether discrepancies are due to model misspecification or simply due to differences in treatment-specific dynamics. Following the prediction-corrected visual predictive check framework, tumor-volume values were therefore rescaled before pooling.

Correction factors were computed within each treatment group and time bin using noise-free prior predictive reference trajectories. Let $a$ index treatment groups, and let $B_b$ denote a time bin. The treatment-specific prior predictive median was defined as
\begin{align*}
m_{a,b}^{\mathrm{pp}}
&=
\operatorname{median}
\left\{
\hat y_i^{(r)}(t):
i\in\mathcal D_{\mathrm{eval}},\,
x_i^{\mathrm{treat}}=a,\,
t\in B_b,\,
R_i^{(r)}(t)=1
\right\},
\end{align*}
where $\hat y_i^{(r)}(t)$ denotes the noise-free prior predictive trajectory. The corresponding pooled reference median was defined as
\begin{align*}
m_{\mathrm{ref},b}^{\mathrm{pp}}
&=
\operatorname{median}
\left\{
\hat y_i^{(r)}(t):
i\in\mathcal D_{\mathrm{eval}},\,
t\in B_b,\,
R_i^{(r)}(t)=1
\right\}.
\end{align*}
Thus, the correction factors were computed using the simulated population that had not dropped out at each time point.

Using zero as the lower bound for tumor volume, the prediction-correction operator for treatment group $a$ and bin $B_b$ was defined as
\begin{align*}
\mathcal P_{a,b}(v)
&=
v
\frac{
m_{\mathrm{ref},b}^{\mathrm{pp}}
}{
m_{a,b}^{\mathrm{pp}}
}.
\end{align*}

Let $b(t)$ denote the time-bin index such that $t\in B_{b(t)}$, and let $a_i=x_i^{\mathrm{treat}}$ denote the treatment group for individual $i$. Observed and simulated tumor volumes were corrected as
\begin{align*}
v_i^{\mathrm{pc}}(t)
&=
\mathcal P_{a_i,b(t)}
\left(
y_i(t)
\right),
\\
\tilde v_i^{(r),\mathrm{pc}}(t)
&=
\mathcal P_{a_i,b(t)}
\left(
\tilde y_i^{(r)}(t)
\right).
\end{align*}

After correction, values from all treatment groups were pooled and the visual predictive check construction described above was applied to $v_i^{\mathrm{pc}}(t)$ and $\tilde v_i^{(r),\mathrm{pc}}(t)$. The resulting dropout-aware prediction-corrected visual predictive check assesses whether the model reproduces the longitudinal tumor-volume distribution after accounting for treatment-specific expected dynamics and dropout-induced selection.

\section{Supplementary Results}
\label{sec:supplementary-results}
This supplementary section provides additional results supporting the analyses presented in the main Results section. The material is organized using the same subsection structure as the main Results section. For each experiment, we include additional quantitative summaries or visual diagnostics that were not shown in the main text.

\subsection*{Mechanistic interpretation and decoder comparison}
\label{sec:supp-mechanistic-interpretation}
\begin{table}[htbp]
\centering
\caption{Hybrid-EB-VAE parameter estimates. Parameter estimates are reported as mean [min, max] across five runs.}
\label{tab:supp-hybrid-parameter-estimates}
\begin{tabular}{lcc}
\toprule
Parameter & Estimate & Reference value* \\
\midrule
$k_g$ 
& 0.0830 [0.0820, 0.0844] 
& 0.06 \\

$a_{\mathrm{LEE011}}$ 
& 0.0458 [0.0445, 0.0464] 
& 0.0156 \\

$a_{\mathrm{encorafenib}}$ 
& 0.0490 [0.0479, 0.0502] 
& 0.0138 \\

$a_{\mathrm{binimetinib}}$ 
& 0.0618 [0.0610, 0.0630] 
& 0.051 \\

$a_{\mathrm{LEE011,encorafenib}}$ 
& -0.0050 [-0.0067, -0.0037] 
& -- \\

$a_{\mathrm{LEE011,binimetinib}}$
& 0.0040 [0.0021, 0.0058]
& -- \\
\bottomrule
\end{tabular}
\end{table}

\begin{table}[htbp]
\centering
\caption{Mean net growth rates for monotherapy treatment groups in the Hybrid-EB-VAE model. Net growth rates were calculated using the mean parameter estimates as \(k_g-a_i\).}
\label{tab:supp-hybrid-net-growth}
\begin{tabular}{lcc}
\toprule
Treatment & Estimate & Reference value* \\
\midrule
Untreated
& 0.0830
& 0.0600 \\

LEE011
& 0.0372
& 0.0444 \\

Encorafenib
& 0.0340
& 0.0462 \\

Binimetinib
& 0.0212
& 0.0090 \\
\bottomrule
\end{tabular}
\end{table}

\FloatBarrier

\subsection*{Population prediction and treatment generalization}
\label{sec:supp-population-treatment-generalization}

\begin{figure}[!htbp]
  \centering

  \includegraphics[width=1\textwidth]{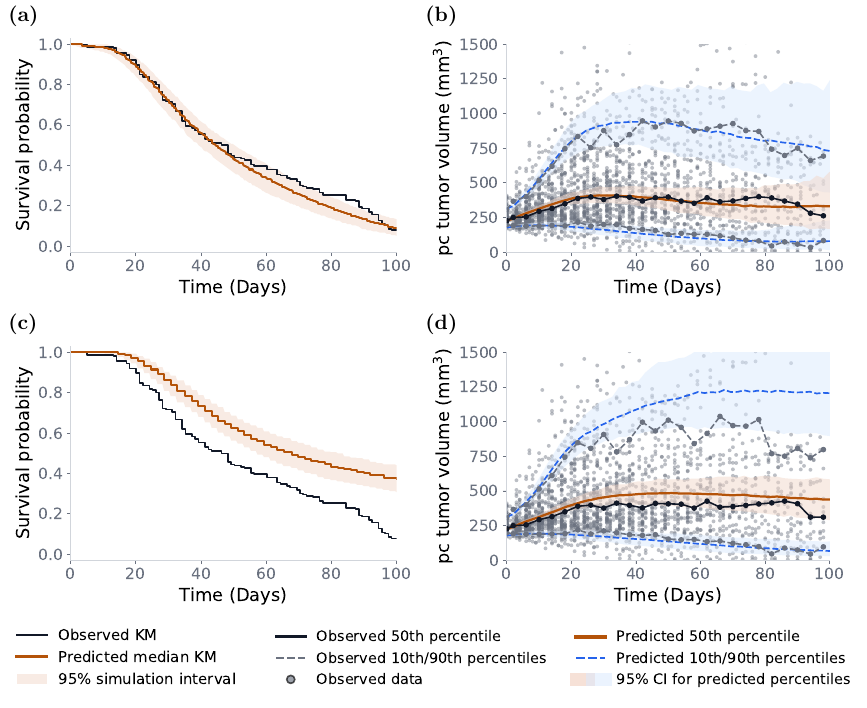}

  \caption{(left) KM-VPCs (right) pcVPCs for one representative cross-validation split of the breast cancer dataset with test individuals pooled across treatment groups. Observed percentiles (not used for training) are shown in black/gray and predicted percentiles in blue/orange, with shaded regions indicating 95\% confidence intervals for the predicted percentiles.
(a) KM-VPC using the learned survival-model dropout mechanism.
(b) pcVPC using the learned survival-model dropout mechanism.
(c) KM-VPC using naive threshold-based dropout.
(d) pcVPC using naive threshold-based dropout.
The KM-VPC legend applies to panels (a,c), and the pcVPC legend applies to panels (b,d).}
  \label{fig:fold0-dropout-vpc-summary}
\end{figure}

\begin{figure}[!htbp]
    \centering
    \includegraphics[width=0.98\textwidth]{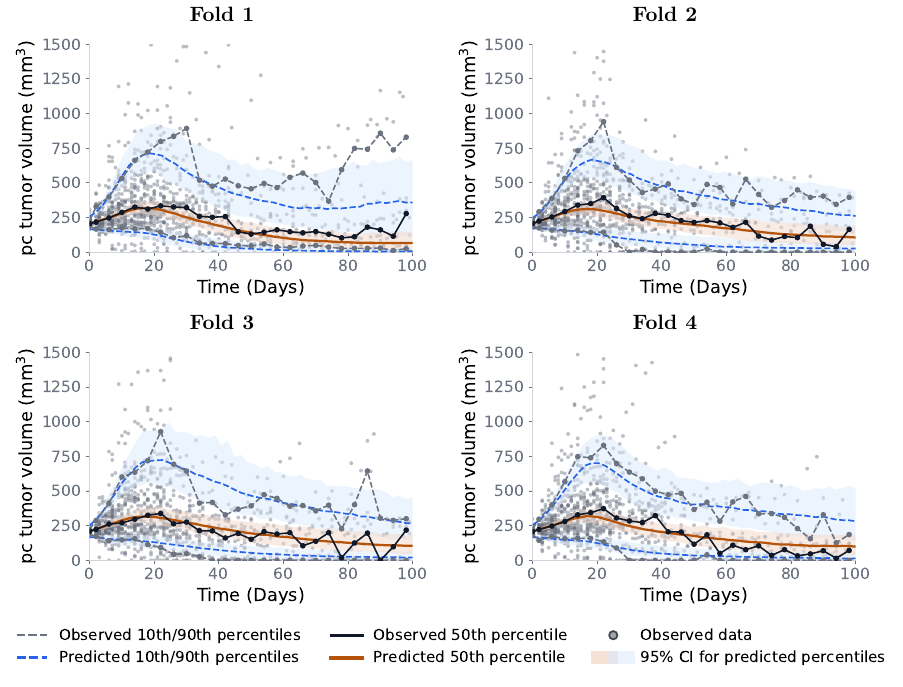}

    \caption{Prediction-corrected visual predictive checks for the remaining cross-validation folds of the cutaneous melanoma dataset. Panels show held-out test individuals pooled across treatment groups for folds 1--4. Observed percentiles are shown in black/gray and predicted percentiles in blue/orange, with shaded regions indicating 95\% confidence intervals for the predicted percentiles. These diagnostics complement the representative fold shown in the main text.}
    \label{fig:supp-cm-fold1234-pcvpc}
\end{figure}

\begin{figure}[!htbp]
    \centering
    \includegraphics[width=0.98\textwidth]{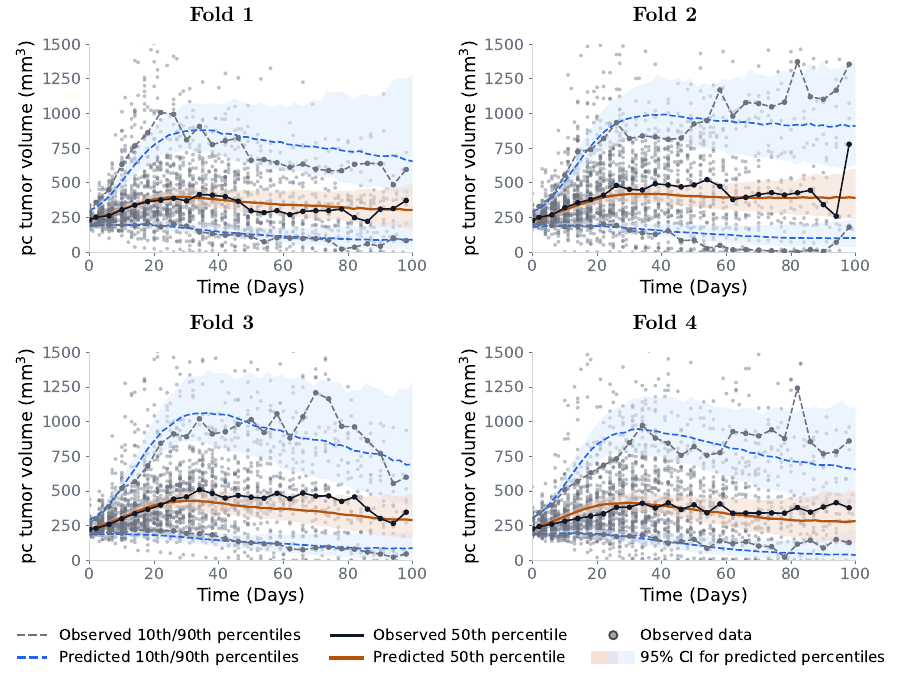}

    \caption{Prediction-corrected visual predictive checks for the remaining cross-validation folds of the breast cancer dataset. Panels show held-out test individuals pooled across treatment groups for folds 1--4. Observed percentiles are shown in black/gray and predicted percentiles in blue/orange, with shaded regions indicating 95\% confidence intervals for the predicted percentiles.}
    \label{fig:supp-bc-fold1234-pcvpc}
\end{figure}

\begin{figure}[!htbp]
    \centering
    \includegraphics[width=0.98\textwidth]{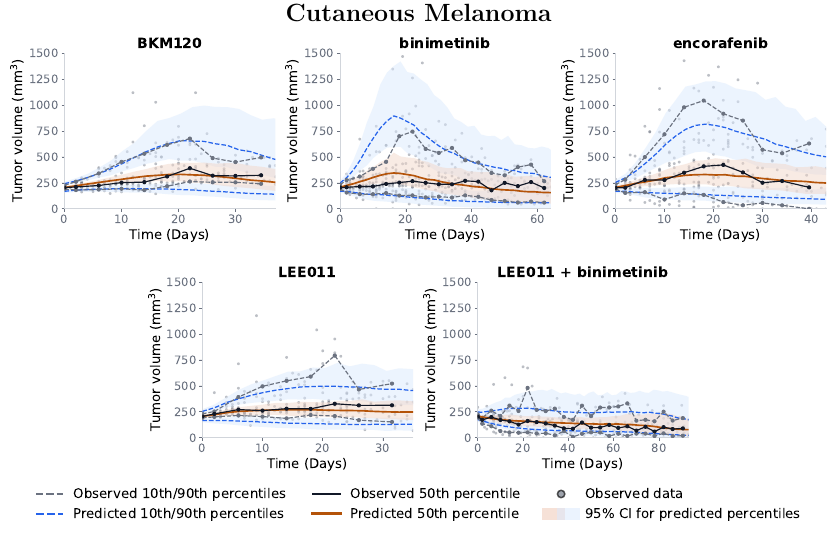}

    \caption{Additional treatment-holdout visual predictive checks for the CM dataset. Each panel evaluates prior predictive performance for a treatment group that was excluded during training. These panels show the remaining held-out treatment groups not included in the main treatment-holdout figure.}
    \label{fig:supp-treatment-holdout-cm-extra}
\end{figure}

\begin{figure}[!htbp]
    \centering
    \includegraphics[width=0.98\textwidth]{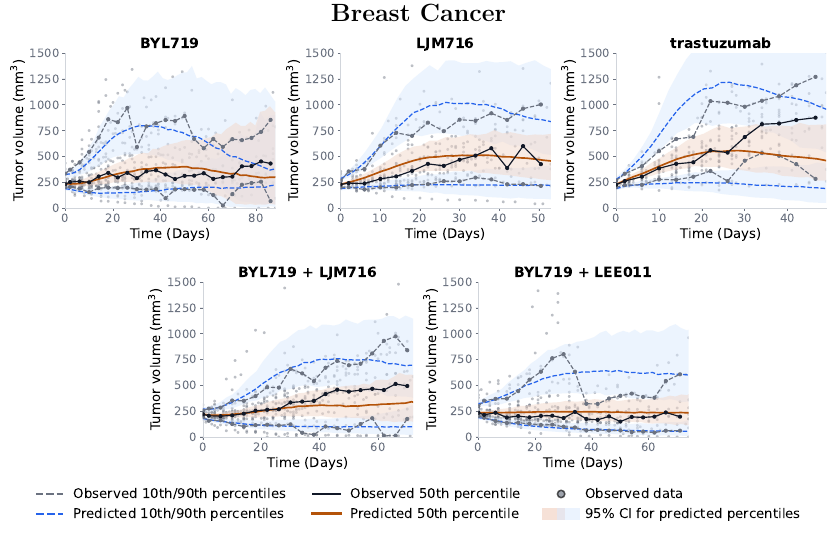}

    \caption{Additional treatment-holdout visual predictive checks for the BC dataset. Each panel evaluates prior predictive performance for a treatment group that was excluded during training. These panels show the remaining held-out treatment groups not included in the main treatment-holdout figure.}
    \label{fig:supp-treatment-holdout-bc-extra}
\end{figure}

\FloatBarrier

\subsection*{Genetic covariates and feature stability analysis}
\label{sec:supp-genetic-covariates-feature-stability}

\begin{table}[htbp]
\centering
\small
\caption{Effect of genetic conditioning on RMSE for the \emph{21-day} datasets. 
\(\Delta\) denotes the absolute RMSE improvement, computed as the treatment-only RMSE minus the RMSE from the model conditioned on genetic covariates. Percent improvement is computed relative to the treatment-only model.}
\label{tab:supp-genetics-rmse-improvement-21day}
\begin{tabular}{@{}llrrrr@{}}
\toprule
Cancer type & Statistic & Without & With & $\Delta$ & $\Delta$ (\%) \\
\midrule
CM & Mean   
& 161.387 & 139.980 & 21.407 & 13.264 \\

CM & Median 
& 106.416 & 93.513  & 12.903 & 12.125 \\

BC & Mean   
& 131.531 & 117.086 & 14.445 & 10.982 \\

BC & Median 
& 79.443 & 71.535 & 7.908 & 9.954 \\
\bottomrule
\end{tabular}
\end{table}

\begin{table}[htbp]
\centering
\small
\caption{
Top genetic indicators from the stability-selection analysis. Selection frequency denotes the percentage of stability selection runs in which the indicator appeared among the top 10 features ranked by permutation-based increase in test KL. The procedure was repeated across five cross-validation folds with 100 runs per fold, using 50 permutations per feature-importance calculation.
}
\label{tab:supp-genetic-stability-selection}
\begin{tabular}{rlllr}
\toprule
Rank & Gene & Indicator & Source & Selection frequency \\
\midrule
1  & NRAS   & Missense mutation             & COSMIC & 65.0\% \\
2  & BRAF   & V600E protein change          & Gao et al. & 62.0\% \\
3  & NRAS   & Q61K protein change           & ChatGPT & 52.4\% \\
4  & SIRPA  & Any protein-changing mutation & Random & 39.4\% \\
5  & NRAS   & Any protein-changing mutation & COSMIC & 32.4\% \\
6  & NF1    & Truncating mutation           & ChatGPT & 26.4\% \\
7  & MDM2   & Amplification                 & COSMIC & 24.0\% \\
8  & BRAF   & Missense mutation             & COSMIC & 23.0\% \\
9  & TBC1D1 & Missense mutation             & Random & 20.4\% \\
10 & SET    & Any protein-changing mutation & Random & 20.4\% \\
\bottomrule
\end{tabular}
\end{table}

\FloatBarrier

\FloatBarrier

\section{Implementation Details}
\label{sec:add-results}
\subsection*{Data Representation and Normalization}

Each PDX mouse was represented as one longitudinal trajectory containing tumor-volume measurements, observation times, treatment information, dosing records, dropout time, and censoring status. Since trajectories have different numbers of observations, minibatches were padded and masks were used so that padded values did not contribute to the training objective.

\paragraph*{Time and dose standardization.}
Time and dose variables were standardized using constants computed from the training split. Observation times, dosing times, and dropout times were scaled by the maximum training follow-up time, while dose amounts were scaled by the maximum training dose. The same transformations were then applied to validation and test individuals, so preprocessing did not use information from held-out data.

\paragraph*{Tumor-volume normalization.}
Tumor volumes were normalized relative to each individual's initial tumor volume. This reduces differences in baseline tumor burden and makes the model focus on relative growth or shrinkage after treatment start. Predictions were transformed back to the original tumor-volume scale for evaluation and plotting.

\paragraph*{Encoder normalization.}
For the encoder, we also used a treatment-specific baseline normalization. Tumor observations were divided by the corresponding median-model prediction before being passed to the encoder. This made the encoder input have a more similar scale across treatment groups, so that the encoder could focus on individual deviations from the typical treatment response rather than on treatment-level differences in magnitude.

\paragraph*{Genetic covariate linking.}
For experiments with genetic covariates, each trajectory also retained the underlying PDX model identifier. This identifier was used to link the trajectory to the corresponding binary genetic indicator vector constructed from the genomic annotations associated with that PDX model.

\subsection*{Model Architecture}

\paragraph*{Posterior encoder.}
The posterior encoder represented each individual trajectory as a variable-length sequence of normalized observation pairs $(t,y)$. The sequence was processed using a transformer-based encoder with masking for padded observations, followed by attention pooling to obtain one individual-level representation. In the main Joint-VAE experiments, the encoder used a transformer dimension of 32, three transformer layers, four attention heads, dropout probability 0.1, and hidden dimension 16 in the feed-forward projection and posterior heads. The pooled representation was mapped to the mean and covariance of a Gaussian approximate posterior over the individual latent parameters.

\paragraph*{Treatment conditioning.}
Treatment information entered the model through learned treatment-conditioning networks. In the CM and BC experiments, treatment was represented by an active-drug vector indicating which drugs were present for the individual. This vector was passed through a treatment encoder consisting of a linear skip connection and a shallow nonlinear correction with one SELU hidden layer. The nonlinear correction was initialized close to zero, so treatment effects were initially close to linear but could become nonlinear during training. Treatment-conditioned outputs were used both as shifts to the posterior mean and to define the empirical Bayes prior mean. A separate treatment-conditioning path also scaled the prior covariance, allowing uncertainty in the individual latent parameters to depend on treatment group.

\paragraph*{Genetic conditioning.}
For experiments involving genetic covariates, the same trajectory encoder and decoder structure was retained, but additional genetic-conditioning modules were added to the encoder. The binary genetic indicator vector was projected to a small set of learned genetic programs; in the implemented genetics experiments, six genetic programs were used. Program scores were normalized and passed through a bounded nonlinearity, then multiplied by treatment-dependent gates computed from the active-drug vector. The resulting genetic representation was mapped to the latent-parameter dimension and combined with a learned PDX-model embedding. Genetic conditioning was used to shift the posterior mean, shift the prior mean, and scale the prior covariance.

\paragraph*{Latent dimensionality.}
The dimension of the individual latent parameter vector \(k_i\) varied across experiments. Shorter follow-up experiments generally used one latent parameter, while longer follow-up and treatment-holdout experiments used two latent parameters. This allowed the model capacity to be increased in settings where longer trajectories required more flexible individual-level variation.

\paragraph*{Neural ODE decoder.}
The main Joint-VAE decoder was a neural ODE decoder. Individual latent parameters sampled from the posterior or prior were held fixed during integration and used to condition the latent dynamics. Treatment entered the decoder through a constant dose signal: for each individual, the dosing records were reduced to one treatment input vector containing the active drugs and their scaled dose amounts. This treatment vector was then used as a fixed covariate in the ODE dynamics over the prediction interval. The ODE function used a feed-forward network with hidden dimension 512, SELU activations, and a linear skip path. After solving the ODE on the dense prediction grid, a linear readout mapped the latent state to tumor volume. The decoder also included a positive hazard head for dropout modeling, and cumulative dropout hazard was obtained by integrating the hazard along the predicted trajectory.

\paragraph*{Hybrid decoder.}
For the hybrid decoder comparison, the neural ODE tumor-growth decoder was replaced by an analytic exponential growth decoder. In this model, the individual latent parameter directly controlled the exponential growth rate, while treatment effects were represented through learned global drug-effect parameters.

\paragraph*{Observation model.}
The observation model used trainable Gaussian noise with additive and proportional components. The same noise model was used for the reconstruction likelihood during training and when sampling noisy predictive trajectories.

\paragraph*{Median model.}
A separate median model was trained before the VAE in experiments using baseline normalization. This model used a deterministic treatment-conditioned encoder rather than a variational trajectory encoder. Its role was to learn typical treatment-level trajectories, which were then used to normalize encoder inputs so that the VAE encoder saw trajectories on a more comparable scale across treatment groups.

\subsection*{Training Procedure}

Training was performed separately for each experimental split. For each fold, preprocessing statistics were computed from the training set and then applied to the training, validation, and test sets. The initial-value normalizer was constructed first, after which a median model was trained. The trained median model was then used to construct the median-model encoder normalizer, and the final VAE was trained using this encoder normalization.

\paragraph*{Median model training.}
The median model was trained before the VAE using an \(L_1\) reconstruction loss and did not include the survival/dropout loss. In the main CM and BC experiments, the median model was trained for at most 200 epochs using Adam, with early stopping based on validation loss. Its purpose was to learn treatment-level typical trajectories used for encoder normalization.

\paragraph*{VAE training objective.}
The VAE was trained using posterior samples of the individual latent parameters. For each minibatch, the encoder produced the approximate posterior \(q_{\phi}(k_i \mid y_i,x_i)\) and the empirical Bayes prior \(p_{\psi}(k_i \mid x_i)\). A sample from the posterior was passed through the decoder, and the resulting dense trajectory was interpolated to the observed measurement times. The training objective combined the observation negative log-likelihood, the Gaussian KL divergence between posterior and prior, and the dropout survival loss. The observation term used the trainable Gaussian noise model described above. The survival term used the decoder-predicted cumulative hazard and hazard at the observed dropout or censoring time.

\paragraph*{Validation and checkpointing.}
Validation metrics were evaluated after each epoch and used for checkpoint selection and early stopping. Test metrics were logged during training for diagnostics only and were not used for model selection. During evaluation, posterior means were used instead of posterior samples to reduce Monte Carlo variation. Model checkpoints were selected using validation performance after an initial burn-in period, and each phase was reset to the best validation checkpoint before continuing. For the main VAE phase, validation checkpointing started after 225 epochs and early stopping used patience 100. For the median model, checkpointing started after 75 epochs and early stopping used patience 50.

\paragraph*{Optimization settings.}
For the main non-genetic Joint-VAE experiments, the VAE phase was trained for at most 1200 epochs using Adam. The posterior, decoder, and default parameter groups used learning rate \(10^{-3}\), while the prior mean and prior covariance parameter groups used learning rate \(10^{-2}\). The observation-noise parameters were frozen for the first 150 epochs and then optimized with learning rate \(2\times 10^{-3}\). Learning rates were reduced on plateau, and gradients were clipped to norm 0.5.

\paragraph*{Genetic prior training.}
For experiments with genetic covariates, training used an additional prior-training phase. During the main VAE phase, the genetic prior parameters were kept frozen. After this phase, the posterior and non-genetic prior components were held fixed, and the genetic prior was trained using a KL-only objective. This phase trained the genetics-conditioned prior to match the already learned posterior distribution, using learning rate \(3\times 10^{-4}\), weight decay \(10^{-4}\), and a warmup-cosine learning-rate schedule.

\end{document}